\documentclass{article}
\usepackage{arxiv}
\usepackage{amsmath}
\usepackage{amsfonts}
\usepackage{graphicx}
\usepackage{colortbl}
\usepackage{xcolor}
\usepackage{algorithm}
\usepackage{algpseudocode}
\usepackage{url}

\title{Advancing calibration for stochastic agent-based models in epidemiology with Stein variational inference and Gaussian process surrogates
}
\author{
  Connor Robertson, Cosmin Safta, Jaideep Ray \\
  Sandia National Laboratories \\
  Livermore, CA\\
  \texttt{cjrobe@sandia.gov} \\
   \And
  Nicholson Collier, Jonathan Ozik \\
  Argonne National Laboratory \\
  Chicago, IL\\
}
\date{August 2024}

\begin{document}
\maketitle
\begin{abstract}
Accurate calibration of stochastic agent-based models (ABMs) in epidemiology is crucial to make them useful in public health policy decisions and interventions.
Traditional calibration methods, e.g., Markov Chain Monte Carlo (MCMC), that yield a probability density function for the parameters being calibrated, are often computationally expensive.
When applied to ABMs which are highly parametrized, the calibration process becomes computationally infeasible.
This paper investigates the utility of Stein Variational Inference (SVI) as an alternative calibration technique for stochastic epidemiological ABMs approximated by Gaussian process (GP) surrogates.
SVI leverages gradient information to iteratively update a set of particles in the space of parameters being calibrated, offering potential advantages in scalability and efficiency for high-dimensional ABMs.
The ensemble of particles yields a joint probability density function for the parameters and serves as the calibration.
We compare the performance of SVI and MCMC in calibrating CityCOVID, a stochastic epidemiological ABM, focusing on predictive accuracy and calibration effectiveness.
Our results demonstrate that SVI maintains predictive accuracy and calibration effectiveness comparable to MCMC, making it a viable alternative for complex epidemiological models.
We also present the practical challenges of using a gradient-based calibration such as SVI which include careful tuning of hyperparameters and monitoring of the particle dynamics.
\end{abstract}

\section{Introduction}
\label{sec:inro}

The calibration of agent-based models has presented a formidable challenge in enabling their widespread adoption and utility.
This is particularly pressing in epidemiology, where agent-based models provide the potential for rapid simulation of interventions to manage pandemic-level diseases~\cite{kerr2021_covid_abm,hinch2021_covid_abm,lorig2021_covid_abm,almagor2020_covid_abm,pescarmona2021_covid_abm}.
As the complexity of these models increases, the need for accurate calibration becomes even more pronounced.
Traditional calibration techniques, such as Markov Chain Monte Carlo (MCMC), have been previously employed for uncertainty quantification tasks due to their convergence guarantees.
However, these methods often carry significant computational cost, which can hinder their applicability for more complex models and for real-time decision-making scenarios.
When the ABM to be calibrated has a large number of parameters, the use of MCMC becomes computationally infeasible.
As a result, ABM simulations are often accelerated using surrogate or emulator model approximations and calibrations are made more affordable using techniques such as approximate Bayesian computation (ABC)~\cite{beaumont2019_abc_review}.

In recent years, alternative calibration approaches have emerged from the field of variational inference that promise to improve the efficiency and scalability of model calibration while maintaining some convergence properties~\cite{pinder2020stein_vi_converge}.
One such method is Stein Variational Inference (SVI)~\cite{liu2016_stein}, a technique that uses gradient information to iteratively refine a set of parameter samples or ``particles.'' This approach offers a compelling advantage in terms of scalability and computational efficiency, particularly for high-dimensional models, where traditional sampling methods struggle~\cite{srikrishnan2021_abm_calib_w_many_params}.
By pairing this estimation technique with surrogate modeling of ABMs with Gaussian processes (GPs), SVI presents a fast and novel pathway for calibration.

This paper aims to explore the utility of SVI as an alternative calibration technique for stochastic epidemiological ABMs, specifically focusing on its application to CityCOVID. CityCOVID is a sophisticated ABM designed to simulate the spread of COVID-19 in Chicago~\cite{ozik2021_citycovid,hotton_impact_2022} and was used to support decision making by public health officials. Specifically, we perform a comparative analysis of SVI and MCMC, evaluating their performance in terms of calibration effectiveness and predictive accuracy. Our findings reveal that SVI not only matches the predictive capabilities of MCMC but also offers a more efficient calibration process, thereby positioning it as a viable and scalable alternative to calibration for ABMs.

Despite its advantages, the implementation of SVI is not without challenges. The calibration process requires careful tuning of the hyperparameters and diligent monitoring to ensure optimal performance. In this paper, we also address these practical considerations, providing insight into the nuances of employing gradient-based calibration methods in the context of agent-based modeling.

The remainder of this paper is structured as follows. This introduction concludes with a literature review that outlines the use of stochastic ABMs in epidemiology, existing calibration techniques for stochastic ABMs generally, and a detailed overview of previous applications of SVI.
Section~\ref{sec:formulation} then outlines the calibration problem, describing the CityCOVID model, the construction of its GP surrogate, and the implementation details for both MCMC and SVI.
In Section~\ref{sec:calibration}, we describe the setup and results of calibration for both methods.
Section~\ref{sec:results} discusses the results of our experiment, showcasing the predictive accuracy and calibration effectiveness of SVI compared to MCMC.
Finally, Section~\ref{sec:conclusion} concludes with a summary of the experiment results and provides directions for future research.

\subsection{Literature Review}
\label{sec:litrev}

ABMs have found a variety of applications across biology~\cite{an2009_bio,zhang2020_bio_ecology,soheilypour2018_bio}, ecology~\cite{mclane2011_ecology,zhang2020_bio_ecology}, economics~\cite{chen2012_economics,axtell2022_economics,hamill2015_economics}, and epidemiology~\cite{hunter2018_epi,marshall2015_epi} and have proven capable in providing predictions of and insight into complex dynamics~\cite{williams2018_complex}. These models simulate the interactions of individual agents, allowing for the exploration of heterogeneous behaviors and local interactions that can lead to emergent phenomena. Several key studies have demonstrated the utility of these models in epidemiological contexts, including the spread of infectious diseases such as influenza, COVID-19, and others.
For instance, Ref.~\cite{li2017_influenza} developed and calibrated an ABM to simulate the transmission dynamics of influenza, highlighting the ability of ABMs to capture the impact of vaccination strategies once tuned.
Similarly, Ref.~\cite{silva2020_covid_social_dist} utilized an ABM to investigate the spread of COVID-19, providing insights into the effectiveness of social distancing measures and demonstrating practical use of ABMs for epidemiological insights.

Effective calibration techniques for these ABMs are essential to ensure that model output aligns with observed data.
Traditional methods, such as grid search and random sampling, have been employed~\cite{quinlan2020_grid}; however, these approaches often struggle with high-dimensional parameter spaces.
More sophisticated techniques, including Bayesian inference and MCMC sampling, have been used because of their theoretical guarantees and uncertainty quantification~\cite{kim2021_mcmc_calib,lux2018_mcmc_calib,grazzini2017_mcmc_calib} but are limited by the usually large computational expense of repeated ABM evaluations.
This is of particular importance for high-dimensional models~\cite{srikrishnan2021_abm_calib_w_many_params}.
Often, this has prompted the use of more heuristic methods such as ABC~\cite{zhang2020_abc_calib,mcculloch2022_abc_calib,ozik2021_citycovid}, genetic algorithms~\cite{d2020_genetic,cess2023_genetic}, or particle swarm optimization~\cite{platt2020_particle_swarm,li2017_influenza}.

Another approach to overcoming the computational challenge of ABM calibration has been addressed via the use of surrogate or emulator models, which approximate the outputs of an ABM using information from previous runs.
Foremost among these approaches are the use of GPs~\cite{kim2021_mcmc_calib, fadikar2018_gp, binois_portfolio_2025}, support vector machines~\cite{angione2022_svm,perumal2022_svm}, bagged and/or boosted machine learning approaches~\cite{ozik_extreme-scale_2018,zhang2020_abc_calib,de2022_rf,kieu2024_rf_nn,perumal2020_rf,lamperti2018_rf,robertson2024_rf,robertson2024_rf_sm}, and neural networks~\cite{cockrell_nested_2020,cess2023_nn,anirudh2022_nn,kieu2024_rf_nn}.
Each of these provide a flexible framework for approximating complex functions, facilitating faster evaluations of the likelihood surface compared to traditional methods.
Recent efforts have also explored the use of spline or equation based modeling via ordinary differential equations as surrogates~\cite{fabiani2024_equation_based,avegliano2023_equation_based,li2017_influenza}.
These provide interpretable and fast simulation alternatives, but require in-depth modeling to either match or coarse-grain the ABM dynamics.
However, the trade-offs between the accuracy of the surrogate and the computational cost of ABM evaluations remains a critical consideration.

The current authors have explored the calibration of epidemiological ABMs, and CityCOVID in particular in Refs.~\cite{robertson2024_rf,robertson2024_rf_sm}.
This was performed using a surrogate model constructed using principal component analysis (to capture the temporal evolution of observables such as daily hospitalizations and deaths) and random forests (to capture their dependence on the ABM parameters being calibrated.
The random forest surrogate also builds in a sensitivity analysis via the Gini impurity, which was used to reduce the number of ABM parameters from 9 key parameters to 4.
These same parameters (which are shown later in Table~\ref{tab:priors}) will be calibrated in this paper, but using a GP surrogate and SVI.
Information on the sensitivity of the GP surrogate approach to these 4 parameters will be included later in Table~\ref{tab:feature_importance}.
The formulation of the inverse problem, the error metrics, figures of merit for checking the predictive skill of the calibration and the training dataset for the surrogates are reused in this work, and are described in detail in Section~\ref{sec:ip} and Section~\ref{sec:surr}; the results from our previous work will be compared with the results from SVI, in Section~\ref{sec:results}.

Outside of ABMs, SVI has emerged as a promising alternative for parameter estimation~\cite{liu2016_stein,biswas2023_stein_calib,morelli2023_stein_calib}.
SVI leverages gradient information to iteratively update several potential parameter samples or ``particles,'' enabling efficient exploration of the parameter space. This method has shown potential for reducing the computational burden of parameter estimation in high-dimensional settings.
This is particularly valuable for ABMs, which often require high-dimensional parameter spaces in order to adequately capture complex dynamics.

Despite the growing interest in SVI, we are unaware of its application to calibration of ABMs and there is limited literature directly comparing its performance to MCMC sampling~\cite{gebraad2021_stein_vs_mcmc}.

\section{Formulation}
\label{sec:formulation}

In order to understand the modeling context, we first outline the parametrization and calibration of the CityCOVID ABM along with our surrogate formulation of it. This includes identifying key parameters to retain for surrogate training and calibration.
We then outline the inverse problem formulation used to determine optimal ABM parameters.

\subsection{Agent-based and surrogate modeling}
\label{sec:models}

CityCOVID is a stochastic agent-based model designed to simulate the spread of COVID-19 in the greater Chicago area~\cite{ozik2021_citycovid,hotton_impact_2022}.
The model captures the dynamics of disease transmission by representing a synthetic population ($\sim$2.6 million individuals) moving between activity locations ($\sim$1.3 million places) based on hourly schedules~\cite{macal_chisim:_2018}.
Each CityCOVID run requires one to specify a random number generator (RNG) seed, which first controls the seeding of the infection (the index cases) in the population and subsequently affects mixing and infection dynamics. Stochasticity due to different RNG seeds results in variation in CityCOVID model outputs. In the calibration procedure described below, we will ignore the stochastic nature of CityCOVID and replace it with its mean behavior for each set of parameter values, computed empirically by averaging over 50 RNG seeds.
Parametrization of the model controls when and how many infections are introduced in the population, the probability of exposure when an individual is co-located with infected individuals, probabilities of engaging in self-protective behaviors and staying at home, changes to infectivity due to isolation, seasonality adjustments, etc.
As such it is well suited to model pandemic interventions such as public health messaging on personal protective behaviors, alternate school or work schedules, and individual isolation.
With this flexibility, CityCOVID has demonstrated predictive accuracy and was used to support decision making by public health officials during the COVID-19 pandemic in Chicago~\cite{ozik2021_citycovid,hotton_impact_2022}.

Previous work constructing surrogates for and calibrating CityCOVID~\cite{robertson2024_rf,robertson2024_rf_sm} has shown that the parameter space can be reduced to four parameters $\vec{\theta}$ whose details are listed in Table~\ref{tab:priors}.

\begin{table}
    \centering
    \begin{tabular}{|c|l|c|}
        \hline
        \rowcolor[gray]{0.8}
        \textbf{Parameter} &
        \textbf{Description} &
        \textbf{Range (min, max)} \\ \hline
        $\theta_1$ &
        Rate of exposure to infected &
        0.046, 0.069 \\ \hline
        $\theta_2$ &
        Time of initial seeding of infections &
        31, 59 \\ \hline
        $\theta_3$ &
        Probability of stay at home &
        0.939, 0.981 \\ \hline
        $\theta_4$ &
        Probability of protective behaviors &
        0.407, 0.492 \\ \hline
    \end{tabular}
    \caption{Retained CityCOVID parameters $\vec{\theta} = [\theta_i]_{i=1}^4$ for surrogate training and calibration as determined in~\cite{robertson2024_rf,robertson2024_rf_sm}. The range of parameters was determined using preliminary runs of CityCOVID~\cite{ozik2021_citycovid}.}
    \label{tab:priors}
\end{table}

To enhance the computational efficiency of the CityCOVID model, we employ a Gaussian process (GP) surrogate $g: \; \vec{\theta} \rightarrow \vec{h},\vec{d}$ where $\vec{h} = \{h_j\}, \vec{d} = \{d_j\}, j =  1  \ldots J$ are the number of \emph{daily} hospitalizations and deaths predicted by the ABM from March 2, 2020 to June 4, 2020, i.e. $J=95$ days.
That is to say, we model the hospitalization and death outputs of CityCOVID as the mean of a Gaussian distribution:
\begin{align}
g(\Theta_*)
&= \mathbb{E}[g_* \mid \Theta,\vec{h},\vec{d},\Theta_*] \nonumber \\
&= K(\Theta_*, \Theta)[K(\Theta, \Theta) + \sigma^2I]^{-1}
\label{eq:gp}
\begin{bmatrix}
    \vec{h} \\
    \vec{d}
\end{bmatrix}
\end{align}
given
\begin{align}
g_* &\sim \mathcal{N}\left(0,
\begin{bmatrix}
K(\Theta, \Theta) + \sigma^2I & K(\Theta, \Theta_*) \\
K(\Theta_*, \Theta) & K(\Theta_*, \Theta_*)
\end{bmatrix}\right)
\end{align}
In this, $g(\Theta_*)$ represent the GP approximation of CityCOVID outputted hospitalizations and deaths for a test set of CityCOVID parameters $\vec{\theta}$ as described in Table~\ref{tab:priors} where $\Theta$ and $\Theta_*$ represent ``training'' and ``testing'' sets respectively.
$K(*,*)$ is a kernel function and $\sigma^2$ represents a ``nugget'' which helps account for noisy training CityCOVID observations $\vec{h}$ and $\vec{d}$.
The testing set $\Theta_*$ is most easily thought of as a set of new parameters at which the surrogate GP model should be evaluated.

The formulation in Equation~\ref{eq:gp} can be reduced to a basis-function like representation written as:
\begin{align}
g(\theta_*)
    &=
    \sum_{i=1}^n
    \alpha_i K(\theta_i, \theta_*), \nonumber \\
    \vec{\alpha} &= (K(\Theta, \Theta) + \sigma^2I)^{-1}
    \begin{bmatrix}
        \vec{h} \\
        \vec{d}
    \end{bmatrix}
    \label{eq:gp_basis}
\end{align}
where $n$ is the number of training samples in $\Theta$ and $\theta_*$ is a test point.
Thus, the GP surrogate can be intuitively considered as a basis representation of the CityCOVID outputs using the kernel function $K$ as a basis function and coefficients related to the kernel-based proximity of the training and test points.
However, this formulation also highlights the shortcoming of a GP surrogate, namely that $\mathbf{\alpha}$ carries the entire training dataset $\Theta$ for any evaluations at a new point $\theta_*$.

In order to tailor the formulation in Equation~\ref{eq:gp_basis} to the specific attributes of CityCOVID hospitalizations and deaths, several hyper-parameters were selected including: the kernel function and its corresponding parameters as well as the nugget parameter $\sigma^2$.
After a process of cross-validation which varied several independent sets of training and test data, the Laplacian kernel function was selected as the most effective.
Given this kernel, the length scale $l$ was selected via gradient based optimization and the kernel nugget $\sigma^2$ was found via a brute force sweep over a range between 0.001 and 1.
The GP surrogate is then validated by comparing its predictions against the outputs of the full CityCOVID model using metrics such as mean squared error (MSE) and R-squared values.


The GP provides a probabilistic framework that captures the underlying relationships between the input ABM parameters and the outputted hospitalizations and deaths at a fraction of the cost of running a full ABM simulation. This validation process ensures that the GP surrogate accurately reflects the behavior of the original model, allowing for efficient exploration of the parameter space during the calibration process.

Though the use of GP surrogates is common for ABMs, one notable detail is the lack of inclusion of time as an input parameter to the GP.
This was found to allow for a more scalable GP without a loss of accuracy. This formulation is appropriate in the calibration setting which uses a fixed set of time comparison points but is not suitable for emulations which require extrapolation in time. This can be equivalently considered as constructing a scalar GP at each specified time point with the caveat that the kernel length scale $l$ is optimized simultaneously across all time. Implementation of this formulation was performed using the GP regressor from the Python package \texttt{scikit-learn}~\cite{pedregosa2011scikit-learn}.

\subsection{Inverse problem}
\label{sec:ip}

There are multiple approaches to the calibration or inverse problem of determining the optimal ABM parameters or a proposed distribution of values $\vec{\theta}^*$.
Previous work~\cite{robertson2024_rf,robertson2024_rf_sm} has demonstrated the utility of a Bayesian inference approach with a surrogate, which will be briefly described.

\textbf{Bayesian inference:}
Bayesian inference allows for control over posterior form and provides some convergence guarantees.
In order to facilitate sampling for this approach, observations of the daily counts of hospitalizations and deaths $\vec{h}^\circ = \{h_j^\circ\},\vec{d}^\circ = \{d_j^\circ\}, j = 1 \ldots J$ are used due to their more stationary nature (as compared to census values).
The discrepancy between these observations and the model prediction (either the ABM or the GP surrogate) is assumed to have a zero-mean Gaussian distribution.
\begin{align}
\begin{bmatrix}
h_j^\circ \\
d_j^\circ
\end{bmatrix}
&=
\vec{g}_j(\vec{\theta})
+
\begin{bmatrix}
\epsilon_h \\
\epsilon_d
\end{bmatrix} \nonumber \\
\epsilon_h &\sim \mathcal{N}(0, \sigma_h^2),\quad
\epsilon_d \sim \mathcal{N}(0, \sigma_d^2).
\end{align}
The errors of hospitalizations and deaths are assumed to be independent of one another as the reporting mechanisms for the data are distinct and involve different testing and analysis procedures.
Under these assumptions, the likelihood is given by:
\begin{align}
    \mathcal{L}(\vec{h}^\circ,\vec{d}^\circ\mid \vec{\theta}) = \frac{1}{(2\pi \sigma_h\sigma_d)^n}\exp\left[-\frac{1}{2}\left(\frac{S_h}{\sigma_h^2} + \frac{S_d}{\sigma_d^2}\right)\right]
    \label{eq:likelihood}
\end{align}
where
\begin{align}
    S_h &= ||\vec{h}^\circ - \vec{g}_h(\vec{\theta})||_2^2 \nonumber\\
    S_d &= ||\vec{d}^\circ - \vec{g}_d(\vec{\theta})||_2^2.
\end{align}
As described in~\cite{robertson2024_rf,robertson2024_rf_sm}, the prior distributions of $\vec{\theta}$ are represented as uniform distributions within specified realistic parameter ranges (shown in Table~\ref{tab:priors}) and the prior distributions of $\sigma_h^2$ and $\sigma_d^2$ are given as non-informative inverse Gamma distributions.
This ultimately yields a posterior form of:
\begin{equation}
    P(\vec{\theta} \mid \vec{h}^\circ,\vec{d}^\circ) \propto
    \frac{1}{(2\pi \sigma_h\sigma_d)^n}\exp\left[-\frac{1}{2}\left(\frac{S_h}{\sigma_h^2} + \frac{S_d}{\sigma_d^2}\right)\right]
    \pi(\vec{\theta})\pi(\sigma_h)\pi(\sigma_d)
    \label{eq:posterior}
\end{equation}
where $\pi(\vec{\theta}),\pi(\sigma_h),\pi(\sigma_d)$ are the prior information for $\vec{\theta}$, $\sigma_h$, and $\sigma_d$ respectively.
The prior $\pi(\vec{\theta})$ is considered non-informative and is represented as a multivariate uniform distribution whose bounds are given in Table~\ref{tab:priors}.
The variance priors $\pi(\sigma_h)$ and $\pi(\sigma_d)$ are assumed to be inverse gamma distributions written as:

\[ \pi(\sigma_h)^{-2} \sim \mathcal{G}\left(\frac{1 + n}{2}, \frac{\zeta_h^2 + S_h}{2} \right) \mbox{\hspace{3mm} and \hspace{3mm}}
   \pi(\sigma_d)^{-2} \sim \mathcal{G}\left(\frac{1 + n}{2}, \frac{\zeta_d^2 + S_d}{2} \right), \]
where $\zeta_h^2 + S_h$ and $\zeta_d^2 + S_d$ are the \emph{rate} parameters of the Gamma ($\mathcal{G}$) distributions and $n$ is the number of observations.

Given this posterior form, delayed rejection Metropolis-Hastings sampling (DRAM)~\cite{smith2024uncertainty} is used to draw samples using the \texttt{pymcmcstat} Python package~\cite{miles2019pymcmcstat}. This procedure results in a Markov chain of samples $\vec{\theta}_k$ where $k=\{1,\ldots,K\}$.
The sequence is checked for stationarity using the method of Raftery and Warnes~\cite{Warnes:2000} from the R package \texttt{mcgibbsit}~\cite{mcgibbsit:Manual} to ensure convergence.
For clarity, this Bayesian inference approach will be referred to as ``DRAM inference'' for the remainder of this paper.

\textbf{Stein Variational Inference:}
In contrast to DRAM inference, SVI instead considers a collection of ``particles'' or samples $\Theta_M = \{\vec{\theta}\}^M_{m=1}$ which are individually iterated according to the gradient of a prescribed posterior form.
The exact iteration procedure is determined by the variational approach of minimizing the Kullback–Leibler (KL) divergence between the empirical distribution $Q(\Theta_m)$ of particle locations and the target posterior form $P(\vec{\theta} | \vec{h}^\circ, \vec{d}^\circ)$ given in Equation~\ref{eq:posterior}:
\begin{equation}
    \underset{Q}{\text{minimize}}\quad\text{KL}(Q || P) = \int Q(x) \log\frac{Q(x)}{P(x)} dx
    \label{eq:kl}
\end{equation}
Given that the KL divergence is hard to minimize directly, the particle based iteration is based on Stein's identity which states:
\begin{equation}
    \mathbb{E}_{\vec{x} \sim P}[\nabla_{\vec{x}} \log P(\vec{x}) \phi(\vec{x}) + \nabla_{\vec{x}}\phi(\vec{x})] = 0
    \label{eq:stein}
\end{equation}
where $\phi(\vec{x})$ is a smooth test function.
This identity is used to construct a potential:
\begin{equation}
    \psi(\vec{\theta}_i) = \frac{1}{M}\sum_{j=1}^Mk(\vec{\theta}_j, \vec{\theta}_i)\nabla_{\vec{\theta}_j}\log P(\vec{\theta}_j \mid \vec{h}^\circ, \vec{d}^\circ) + \nabla_{\vec{\theta}_j}k(\vec{\theta}_j, \vec{\theta}_i)
    \label{eq:stein_potential}
\end{equation}
where $k(\vec{\theta},\vec{\theta}')$ is a kernel function which computes a distance between $\vec{\theta}$ and $\vec{\theta}'$.
We follow the most common kernel approach used with Stein which is to assume a Gaussian kernel $k$~\cite{liu2016_stein}.
The potential $\psi(\vec{\theta})$ can be used with any gradient ascent algorithm to provide an optimal perturbation direction for an individual particle to reduce the KL divergence.

In short, this optimization maximizes the posterior of the observed values while accounting for a ``repulsion'' between particles to avoid collapse on the mode of the posterior distribution.
Specifically, the first term in the sum in Equation~\ref{eq:stein_potential} would facilitate a maximization of the log posterior while the second term in the sum provides a repulsion based on the gradient of the distance between particles.
While SVI does not explicitly assume a mean-field approximation $Q(\Theta_M) = \prod_{\vec{\theta}_i \in \Theta_M}Q(\vec{\theta}_i)$, because the optimal perturbation direction of each particle is individually computed, it is assumed to behave like a mean field approximation.
Thus it is not guaranteed that the particles represent samples from the specified posterior distribution in Equation~\ref{eq:posterior}, but particles are considered to be samples from the variational distribution $Q(\Theta_M)$.

\begin{algorithm}
\caption{Stein Variational Inference (SVI)}
\label{alg:SVI}
\begin{algorithmic}[1]
\Require Target distribution \( P(\vec{\theta}) \), initial particles \( \{\vec{\theta}_i\}_{i=1}^M \), kernel function \( k(\vec{\theta}, \vec{\theta}') \)
\Ensure Updated particles \( \{\vec{\theta}_i\}_{i=1}^M \) approximating \( P(\vec{\theta}) \)

\For{iteration $t = 1$ to $T$}
    \For{each particle $i = 1$ to $M$}
        \State Compute the gradient of the log target density: \( \nabla_{\vec{\theta}_i} \log P(\vec{\theta}_i) \) (calls surrogate function)
    \EndFor

    \For{each particle $i = 1$ to $M$}
        \State Initialize the update direction: \( \phi(\vec{\theta}_i) \gets 0 \)
        \For{each particle $j = 1$ to $M$}
            \State Compute the kernel value: \( k_{ij} \gets k(\vec{\theta}_j, \vec{\theta}_i) \)
            \State Compute the gradient of the kernel: \( \nabla_{\vec{\theta}_j} k_{ij} \gets \nabla_{\vec{\theta}_j} k(\vec{\theta}_j, \vec{\theta}_i) \)
            \State Update the direction:
            \[
            \psi(\vec{\theta}_i) \gets \psi(\vec{\theta}_i) + \frac{1}{M} \left( k_{ij} \nabla_{\vec{\theta}_j} \log p(\vec{\theta}_j) + \nabla_{\vec{\theta}_j} k_{ij} \right)
            \]
        \EndFor
    \EndFor

    \For{each particle $i = 1$ to $M$}
        \State Update the particle \(\vec{\theta}_i\) using selected gradient ascent algorithm (using step size and other hyperparameters)
    \EndFor
\EndFor
\end{algorithmic}
\end{algorithm}

Note that in contrast to the DRAM inference approach, this method also requires gradient information of the posterior, which includes the ABM or surrogate-computed outputs. Thus, the ABM or surrogate must be differentiable which limits the use of most ABMs and many surrogates including random and gradient boosted forests. Though closed form gradients exist for GP surrogates, this work made use of simple finite differences to allow for future generalization to other non-differentiable surrogate approaches. This added negligible computational expense, even in multiple dimensions, due to the speed of evaluation of the GP.

This method provides several potential areas of scalability as compared to DRAM.
First, due to the gradient informed nature of the particle evolution, SVI is inherently more efficient in high dimensional spaces.
Though DRAM does include intelligent sampling and acceptance strategies, new samples include significant randomness making them inefficient at exploring high dimensional spaces.
Alternatively, each sample or particle in SVI is following gradient information for improved efficiency.
Second, the repeated evaluations of the surrogate for particle updates in SVI can be parallelized leading to improved scaling.

\textbf{Comparisons:}
In order to compare the two inference techniques, several measures of predictive accuracy are considered.
These include a posterior predictive test (PPT), the continuous rank probability score (CRPS)~\cite{gneiting2005_crps}, and verification rank histogram (VRH)~\cite{hamill2001rank_verification_hist}.
These can be described as follows:
\begin{description}
    \item[Posterior predictive test (PPT)]
        In essence, this is a comparison of the posterior predictive distributions of each calibration approach.
        These are most easily compared with a mean.
        The posterior predictive distributions consist of CityCOVID hospitalization and death predictions using posterior samples from each respective calibration approach.
        This comparison can be written as:
        \begin{align}
            \text{PPT}(h,d) &= \|\mathbf{E}_{\vec{\theta}}[P(h,d \mid \vec{\theta}_\text{MCMC})] - \mathbf{E}_{\vec{\theta}}[P(h,d \mid \vec{\theta}_\text{SVI})]\|_2
            \label{eq:ppt}
        \end{align}
    \item[Continuous rank probability score (CRPS)]
        This comparison aims to cast the deterministic observed values into a degenerate distribution via indicator functions.
        Once expressed in this form, the cumulative distribution functions of an ensemble of predictions from the posterior predictive distribution are compared with this degenerate distribution. The CRPS is computed for each time $t$ for which we have an observation. This can be written as:
        \begin{equation}
            \text{CRPS}(h,d) = \int_{0}^{\infty} \left(P(x | \vec{\theta}) - \mathbf{1}_{\{x \geq h,d\}}\right)^2 dx,
            \label{eq:crps}
        \end{equation}
        where $\vec{\theta}$ are drawn from the posterior distribution and $P(x | \vec{\theta})$ is the cumulative distribution function.
    \item[Verification rank histogram (VRH)]
        Given a finite number of samples from the posterior predictive distribution, this comparison analyzes where the observations rank in terms of size relative to the predictions.
        This captures how well the predictive posterior bounds the observed values.
        For example, if the histogram is heavily skewed toward smaller values, the observations are consistently the largest values and the predictive posterior is underestimating the true values.
        This is more easily seen visually than written as an equation.
        See Figure~\ref{fig:compare_vrh} for an example.
\end{description}

Additionally, the results of both the DRAM and SVI inference will be compared with the original calibration of CityCOVID which used an iteratively refined version of approximate Bayesian computation (ABC)~\cite{ozik2021_citycovid}.
This approach utilized several rounds of CityCOVID evaluations in which only parameters which yielded hospitalizations and deaths sufficiently close to the observed values were accepted and included in the posterior distribution.

\section{Calibration}
\label{sec:calibration}

Given the formulations in the previous section, the surrogate model can be constructed and used for repeated sampling using both the DRAM inference and SVI approaches.
This includes choosing suitable hyperparameters for the GP surrogate and calibration methods.

\subsection{Surrogate construction}
\label{sec:surr}

In order to provide a fair comparison between the two calibration approaches, they were tested on a dataset compiled for previous work on the surrogate modeling and calibration of CityCOVID~\cite{robertson2024_rf,robertson2024_rf_sm}. This dataset consists of a Halton sequence of length 700 over a four dimensional hypercube of the parameters $\vec{\theta}$. The bounds for each parameter are listed in Table~\ref{tab:priors}. For each parameter set, the hospitalizations and deaths simulated with CityCOVID, averaged over 50 different random replicates, were collected.

\begin{figure}
    \centering
    \includegraphics[width=\linewidth]{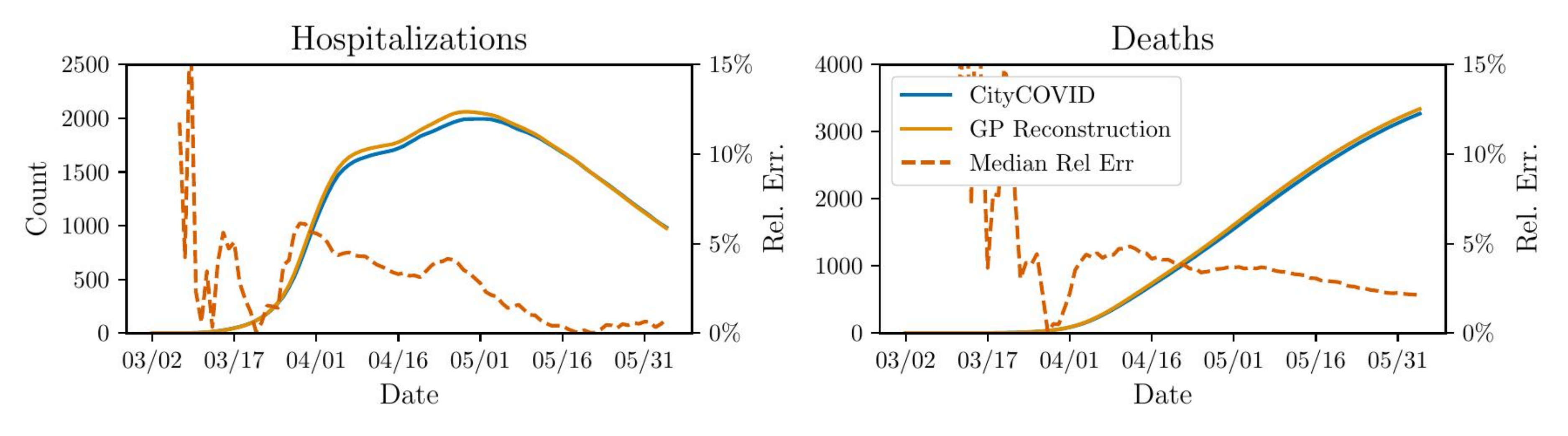}
    \caption{Example hospitalization and death trajectories from the GP surrogate compared with the CityCOVID outputted values for a test parameter set.}
    \label{fig:gp_accuracy_single}
\end{figure}

To test the surrogate accuracy, the dataset was randomly shuffled and split into 5 subsets and cross validation was performed by training on 4 of the subsets while testing on the remaining subset.
The average test error in this cross validation showed a median absolute relative error of 2.5\%.
In this case, the median was used to avoid weighting the relative errors at early times too heavily. These counts at early times are smaller and thus more sensitive to relative errors and less important for predictive accuracy. This can be observed in an example of the GP predicted hospitalizations and deaths compared to the CityCOVID hospitalizations and deaths on one of the hold-out test cases in Figure~\ref{fig:gp_accuracy_single}, where the GP surrogate is shown to closely resemble the CityCOVID outputs.
It can be noted that the largest relative errors occur early in the simulation due to the low occurrences at that time.
For example, early in the simulation a surrogate may predict 2 or 3 deaths when the true value may be 1, which is a relative error of 200-300\%.
Though these early times may be important for analysis, they are not of particular importance in the context of calibration.
We thus consider a median absolute relative error when assessing the surrogate accuracy, which weights the early errors less than a mean.

\begin{figure}
    \centering
    \includegraphics[width=0.45\linewidth]{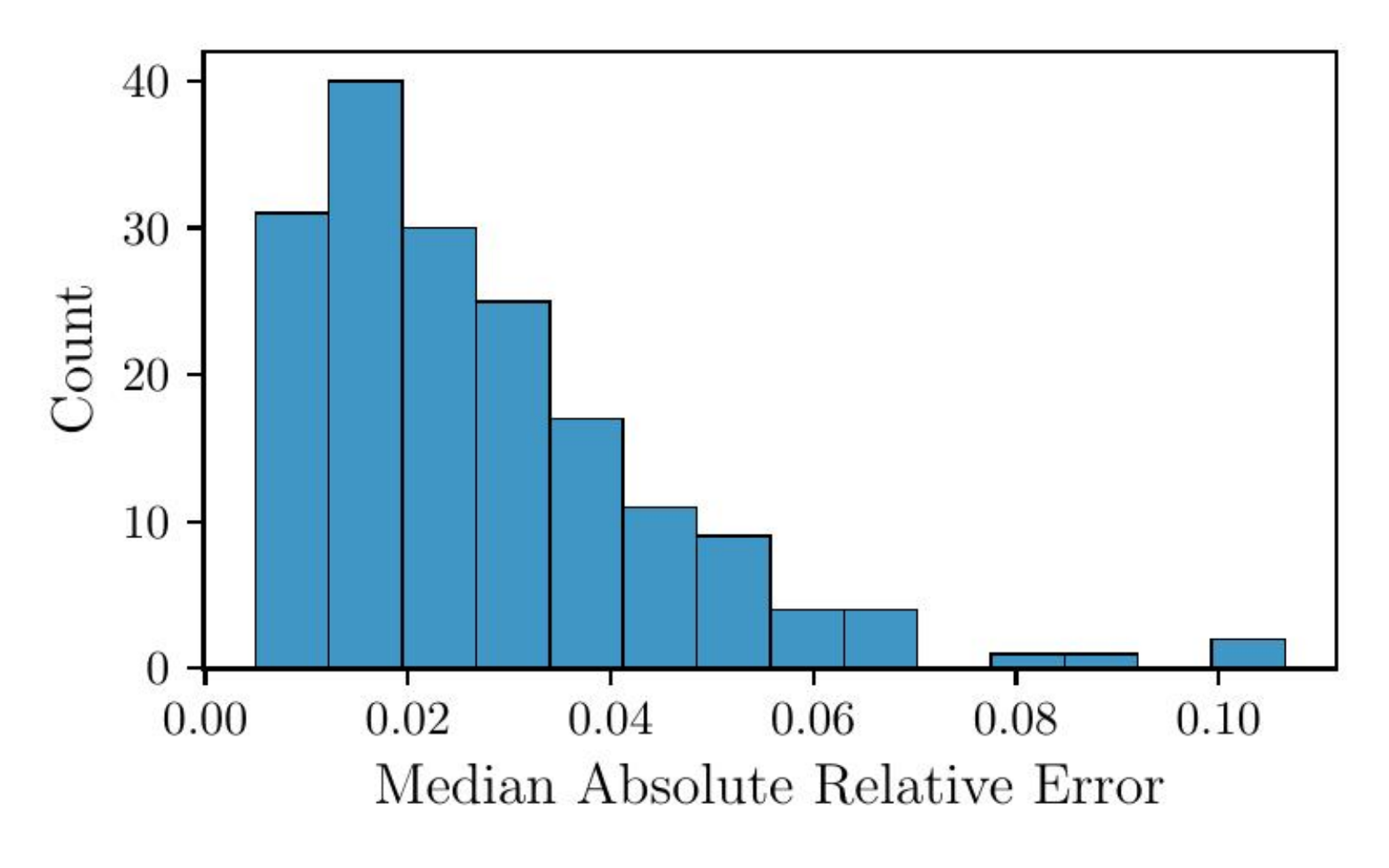}
    \caption{Histogram of median absolute relative error for the GP surrogate applied to all test parameter sets (20\% of dataset).}
    \label{fig:gp_accuracy_hist}
\end{figure}

In order to get a sense of the distribution of absolute relative errors across a range of test parameter values, the dataset was split into training (80\%) and test (20\%) sets.
After training the GP with the relevant data, it was applied to the held out test set and the absolute relative errors were computed.
These errors are shown in the histogram in Figure~\ref{fig:gp_accuracy_hist}.
This histogram shows that the surrogate was able to reproduce the majority of the held out test data with a median absolute relative error of less than 5\%.
There are a few outliers (up to 10\%), which generally consisted of cases of very low or very high hospitalizations or deaths.
This level of accuracy does bring up a concern that the GP may be over-fit or incapable of prediction in other times.
However, as the objective of this work is calibration, the comparison times are fixed and the space filling Halton sequence samples used to train the GP allow for some robustness to overfitting.

In order to assess the global sensitivity of the GP surrogate, the GP was assessed for permutation importance as well as first and total Sobol indices.
A brief description of these global sensitivity analyses are as follows:
\begin{description}
    \item[Permutation importance]
    A measure of accuracy of the GP when the input data of a parameter is shuffled.
    Severely reduced accuracy when a single parameter is shuffled indicates reliance on information from that parameter.
    \item[Sobol (first)]
    A measure of the variance of the output of the GP across variations in a parameter.
    Significant changes in output from adjustments to a single parameter indicates reliance on information from that parameter.
    \item[Sobol (total)]
    A measure of the variance of the output of the GP across variations in a parameter and that parameter in combination with others.
    Significant changes in output from adjustments to a single parameter indicates reliance on information from that parameter.
    This total form also attempts to include nonlinear interactions with other parameters.
\end{description}

\begin{table}[h]
    \centering
    \begin{tabular}{|l|c|c|c|}
        \hline
        \rowcolor[gray]{0.8}
        \textbf{Feature} &
        \textbf{Permutation Importance} &
        \textbf{Sobol (first)} &
        \textbf{Sobol (total)}
        \\ \hline
        $\theta_1$: Rate of exposure to infected &
        0.91 &
        0.50 &
        0.53
        \\ \hline
        $\theta_2$: Time of initial seeding of infections &
        0.56 &
        0.29 &
        0.31
        \\ \hline
        $\theta_3$: Probability of stay at home &
        0.21 &
        0.09 &
        0.11
        \\ \hline
        $\theta_4$: Probability of protective behaviors &
        0.05 &
        0.02 &
        0.03
        \\ \hline
    \end{tabular}
    \caption{Gaussian process feature importance metrics for CityCOVID parameters from~\cite{robertson2024_rf,robertson2024_rf_sm}}
    \label{tab:feature_importance}
\end{table}
Similarly to the previous surrogate modeling of CityCOVID in~\cite{robertson2024_rf,robertson2024_rf_sm}, the rate of exposure to infected and the time of initial seeding were by far the two most important parameters.
The other parameters played some role but were relatively minor in comparison.

\subsection{Calibration details}
\label{sec:calib}

After training the GP surrogate, the posterior distribution formulated in Equation~\ref{eq:posterior} was sampled using the DRAM approach for $K=200,000$ samples.
Convergence tests were performed for this sampling using the \texttt{mcgibbsit} R package which checks for stationarity as a stopping criterion, using the method by Raftery and Warnes~\cite{Warnes:2000}, which builds on an older method by Raftery and Lewis~\cite{92rl2a}.
The tests revealed that 145,000 samples were needed and thus that the sampling had indeed converged.

The gradient-based optimization of the SVI was performed using the Adam optimizer~\cite{kingma2014adam}.
As is common with most gradient-based optimizers popularized by deep learning, this approach adapts the learning rate via ``momentum'' which is determined from exponential averages of the gradients.
This approach is designed to minimize oscillations around a global minimum while skipping past local minima.
However, this step-complexity requires additional hyperparameter tuning of the initial learning rate and the rate of exponential decay of the first and second moments of the gradients.
The selection of these parameters is further complicated by the number of particles used in the SVI calibration.
Specifically, more particles will increase the probability of exploding inter-particle repulsion when particles are pushed too near one another and the initial learning rate and exponential decay must be adjusted accordingly.
The initial learning rate and decay then affect the number of gradient steps required to converge.
For simplicity, we assumed standard values of 0.9 and 0.999 for the exponential decay rates of the first and second gradient moments and a Gaussian kernel with automatic bandwidth selection~\cite{liu2016_stein} for SVI.
The automatic bandwidth is set to balance the contribution of individual particles gradients with those of the other particles.
Thus, the list of tunable hyperparameters was reduced to the number of particles, the initial learning rate, and the number of gradient steps.

\begin{figure}
    \centering
    \includegraphics[width=.6\linewidth]{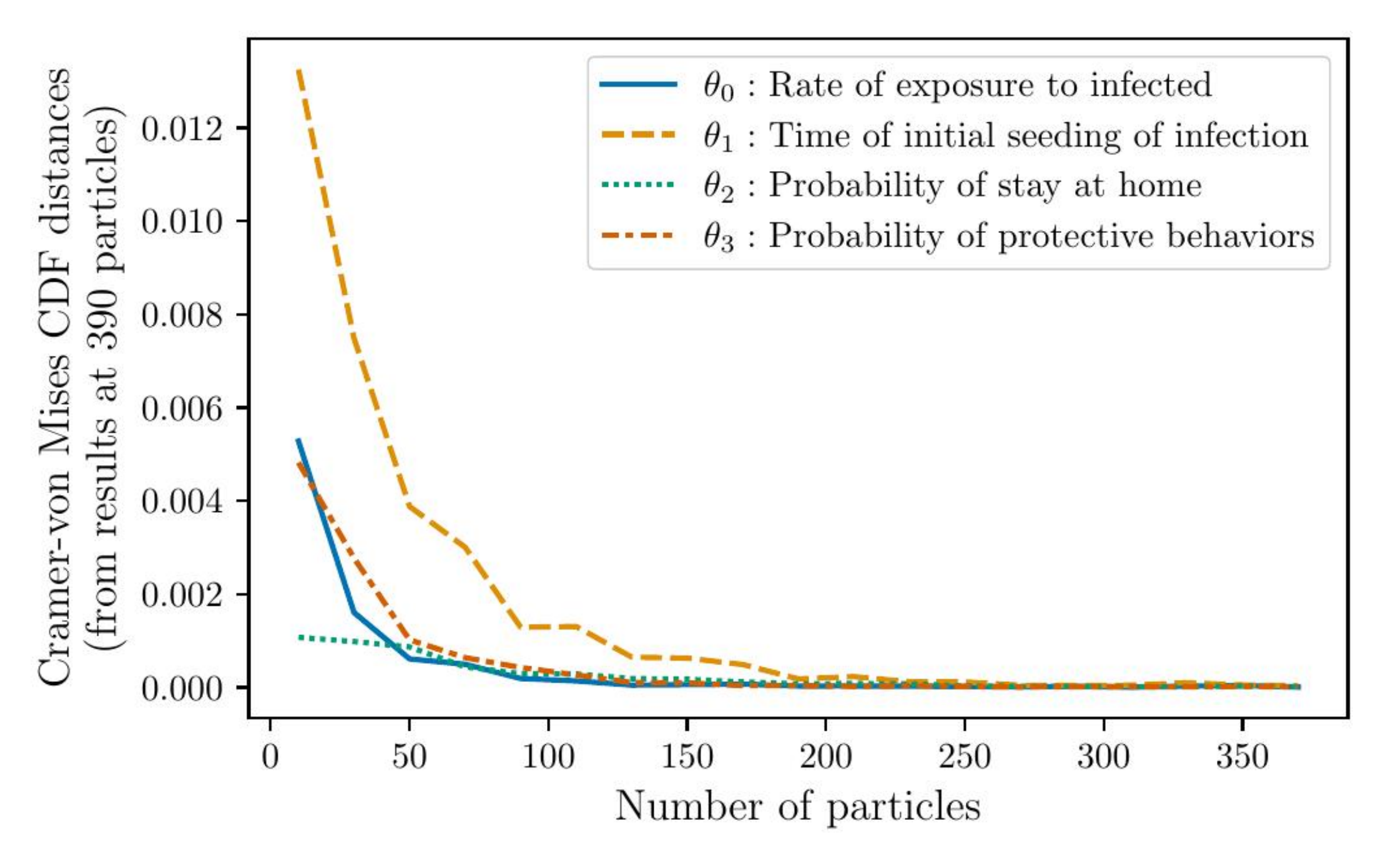}
    \caption{Self-consistency of SVI for increasing number of particles showing the asymptotically decreasing Cramer-von Moises criterion of posterior distributions to the distribution using 300 particles. Each posterior distribution is a combination of 10 distributions using the same number of particles but different initialization.}
    \label{fig:self-consistency}
\end{figure}

The number of particles was first determined by looking for self-consistency.
That is to say, an ``approximately true'' posterior distribution was computed by performing the calibration with a ``large'' number of particles (400).
The calibration was then performed using successively smaller numbers of particles and the resulting marginal posterior distributions were compared with the approximately true marginal posterior using empirical distribution functions and the Cramer-von Mises criterion, a mean-squared error-like comparison of cumulative distribution functions.
Unlike distance measures such as Kolmogorov-Smirnov or Wasserstein distances which use absolute differences and thus emphasize matching the tails of the marginal posteriors, the Cramer-von Mises criterion focuses on matching the distribution modes~\cite{darling1957kolmogorov}.
For each calibration, a very small learning rate and many gradient steps were used to ensure convergence without blow up from inter-particle repulsion.
The results of this self-consistency test are shown in Figure~\ref{fig:self-consistency}.
From the results demonstrated in this figure we selected the number of particles to be 200 as the self-consistency was only marginally improved with more particles.

\begin{figure}
    \centering
    \includegraphics[width=\linewidth]{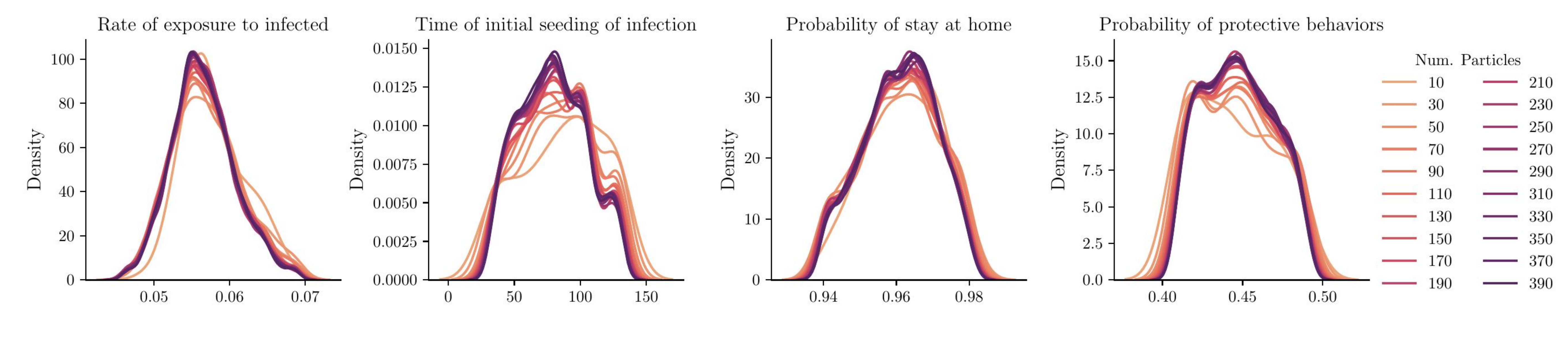}
    \caption{Comparison of posterior distributions yielded by SVI for varied numbers of particles. Each posterior distribution is a combination of 10 distributions using the same number of particles but different initialization.}
    \label{fig:distribution_convergence}
\end{figure}

In order to understand the impact of selecting the number of particles on the posterior distributions compared with the Cramer-von Mises criterion, Figure~\ref{fig:distribution_convergence} shows the posterior distributions used in the self consistency test.
As can be observed, the general qualitative behavior of the distributions is consistent, however the modes can be shifted when too few particles are used in the calibration.

After some manual experimentation using 200 particles, the learning rate was set to 0.001 with 10,000 gradient steps.
These parameters balanced attaining reasonably quick convergence while avoiding blow up from inter-particle repulsion.
A single calibration made up of 10 random initializations took only about 45 minutes on a laptop computer.

\begin{figure}
    \centering
    \includegraphics[width=\linewidth]{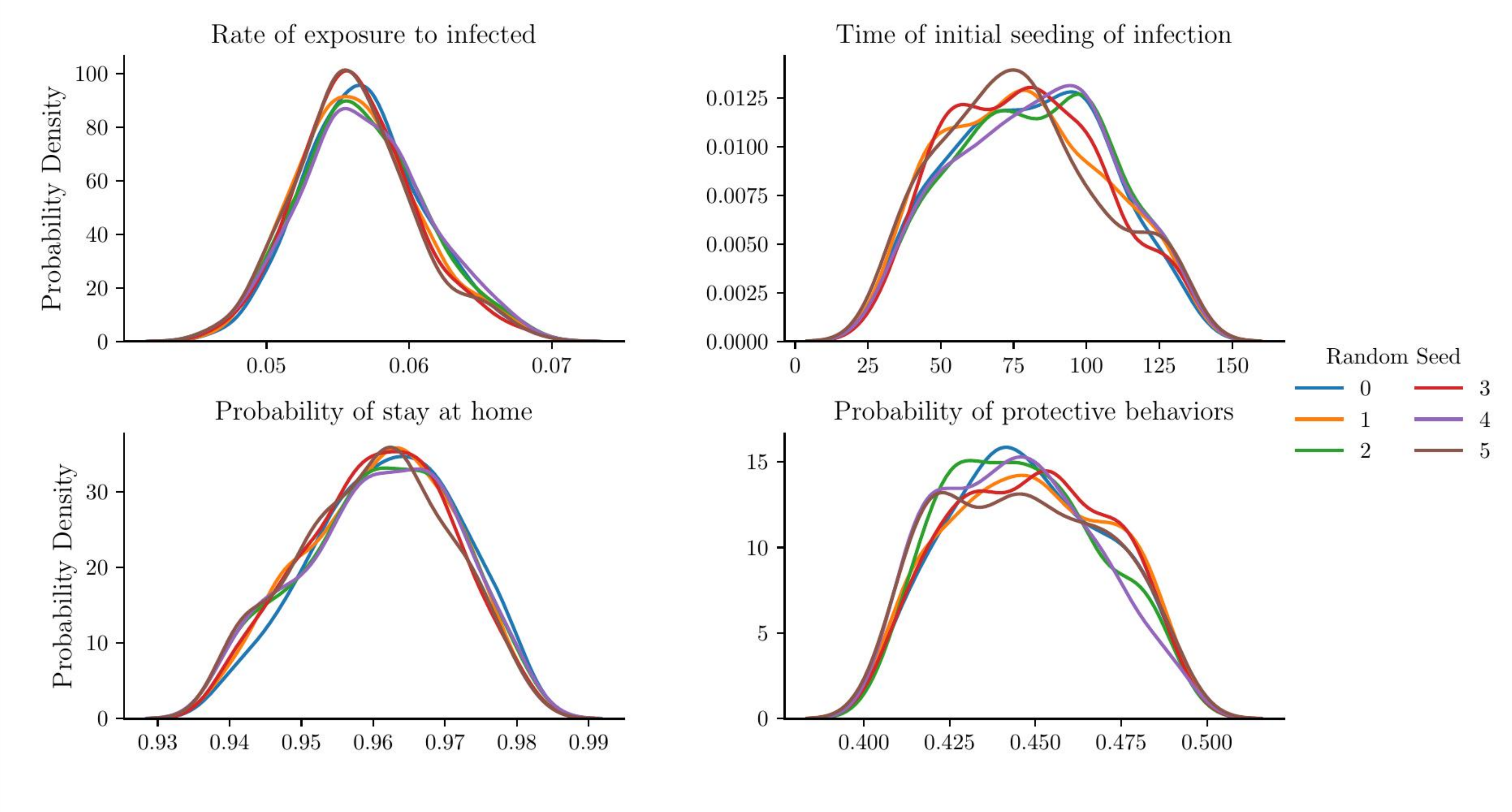}
    \caption{Kernel density estimate representation of posterior distributions from SVI calibration using 200 particles and different particle initializations based on random seed.}
    \label{fig:seed_convergence}
\end{figure}

Another consideration necessary for the SVI optimization is the initialization of the particles. Depending on the density and random initialization of these particles, there are subtle differences in the resulting distribution. Figure~\ref{fig:seed_convergence} demonstrates these differences in the marginal distributions of the parameters following a calibration run of 200 particles.
To ameliorate theses differences, 10 random seeds were used to provide varied initializations for each SVI run.
The resulting samples from each run were then combined into a superset of samples which serve as draws from the posterior.
This approach is used for all SVI runs in this paper.

The iterative updates of the particles in the final optimization procedure can be seen marginally in Supplementary Video 1 along with the convergence of the log posterior probability as defined in Equation~\ref{eq:posterior} across all particles and their respective daily hospitalization and death trajectories.

\section{Results \& Discussion}
\label{sec:results}

\begin{figure}
    \centering
    \includegraphics[width=.9\linewidth]{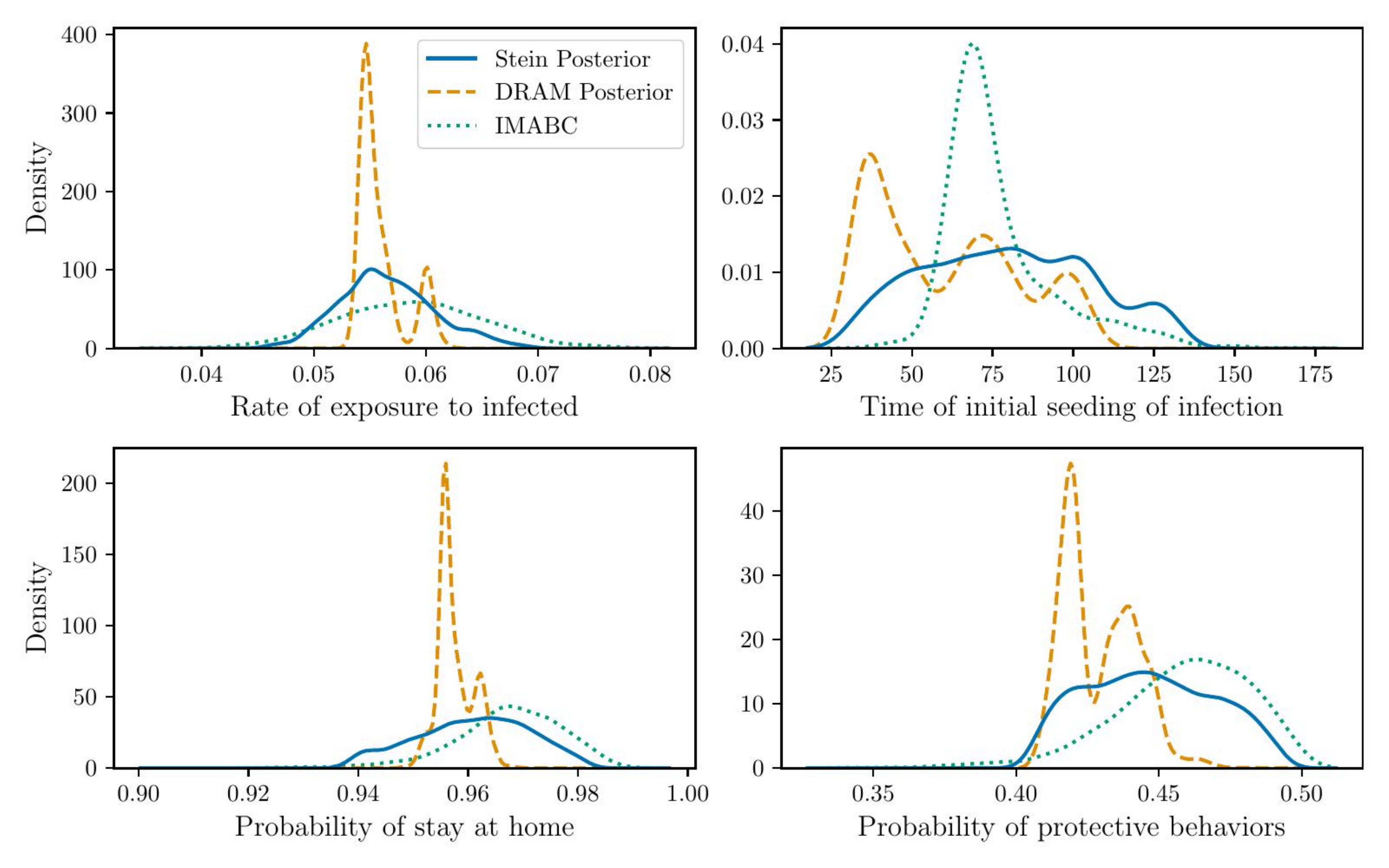}
    \caption{Comparison of posterior distributions from SVI, sampling of the Bayes posterior using DRAM, and the original CityCOVID calibration using IMABC.}
    \label{fig:posterior_compare}
\end{figure}

\begin{figure}
    \centering
    \includegraphics[width=\linewidth]{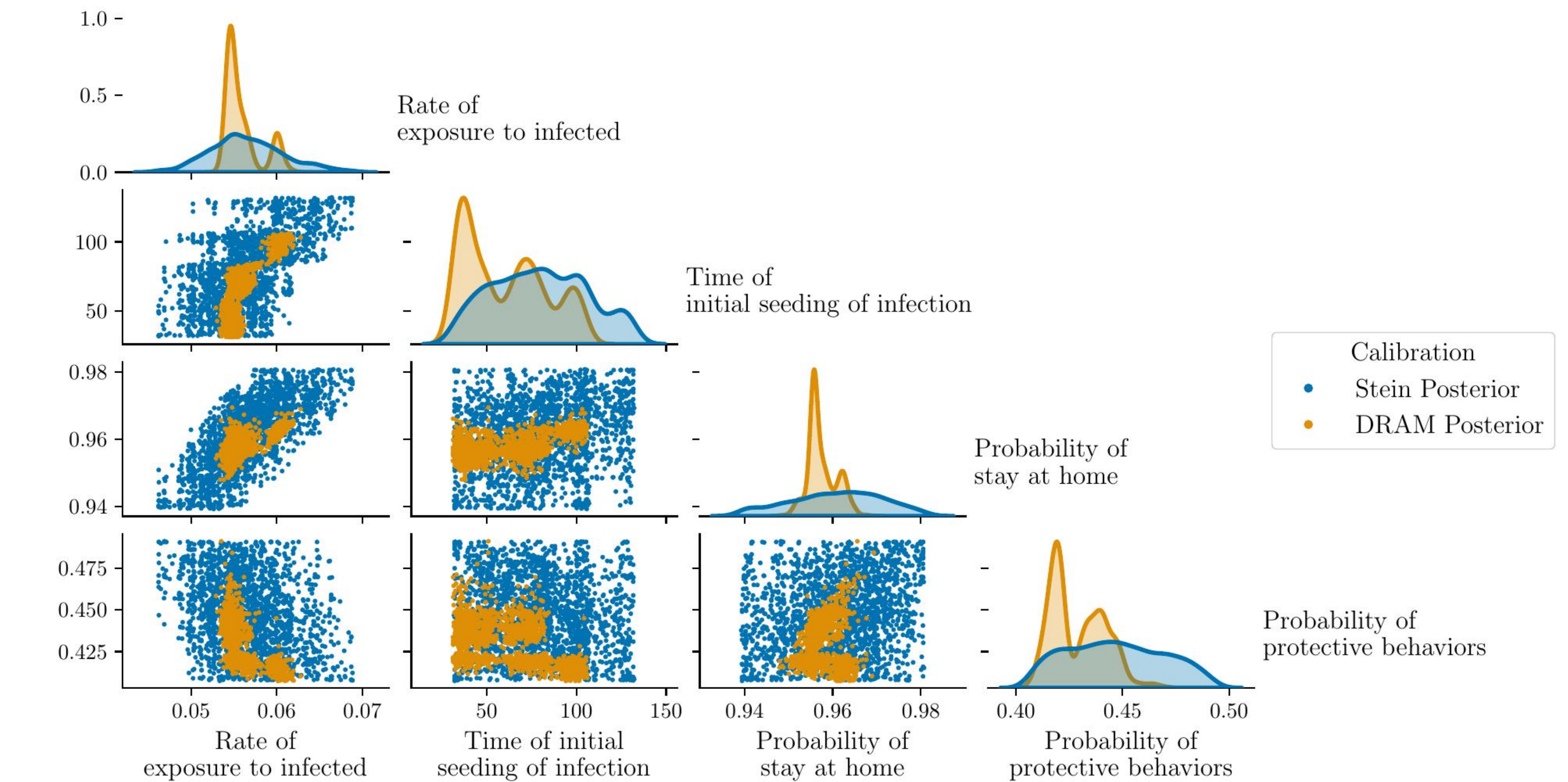}
    \caption{1-D marginal posterior distributions via kernel density estimate (KDE) and pairwise posterior samples from the Bayesian calibration using DRAM and the SVI calibration.}
    \label{fig:pairwise_compare}
\end{figure}

The resulting posterior distributions via DRAM sampling, the SVI, and the original approximate Bayesian calibration (ABC) performed for CityCOVID are shown in Figure~\ref{fig:posterior_compare}.
Pairwise plots of the DRAM and SVI posterior samples are shown in Figure~\ref{fig:pairwise_compare}.
Note that the SVI results in Figure~\ref{fig:pairwise_compare} include 200 samples (particles) each with 10 random initializations yielding a total of 2000 scatter points.

Immediately noticeable is the multi-modal nature of the posteriors from the DRAM results when compared with the SVI and original ABC approach in~\cite{ozik2021_citycovid} which was discussed at the end of Section~\ref{sec:ip}.
This behavior is also not present in the previous random forest surrogate-based calibration from ~\cite{robertson2024_rf,robertson2024_rf_sm} and we thus ascribe it to the flexibility and improved accuracy of the GP surrogate used here.
While this multimodality may be initially unsettling, it is easily explained by the discrete nature of the $\theta_2 =$ ``Time of initial seeding of infection" parameter.
Though this parameter is represented as continuous (in units of hours), infections are introduced into the model population at the beginning of the day corresponding to the 24-hour span that includes the initial seeding time.
Thus, the DRAM posteriors indicate that a reasonable parameter selection would be to let $\theta_2\approx$ 48,72, or 96 hours.

Increasing the time of initial seeding of infection intuitively necessitates an increased rate of exposure in order to match the initial increases of hospitalizations and deaths in the observed data.
This exact relationship can be observed in the pairwise scatter plots between $\theta_1$ and $\theta_2$ in Figure~\ref{fig:pairwise_compare}.
Note that in this plot the largest posterior mode in the rate of exposure to infected ($\theta_1$) also correlates with a later time of initial seeding of infection ($\theta_2$).

Increasing the rate of exposure to infection ($\theta_1$) is also intuitively tied to an increase in the probability of stay at home ($\theta_3$) as this would cause an initial uptick in hospitalizations and deaths that would match the observed data.
This correlation is also represented in the pairwise scatter plot of Figure~\ref{fig:pairwise_compare}.

Though the marginal posteriors from the SVI calibration observed in Figure~\ref{fig:posterior_compare} are much broader than the DRAM calibration, they share several qualitative characteristics from the DRAM posteriors.
Firstly, the marginal posterior for the time of initial seeding of infection parameter has a slight multi-modal shape which shows some preference for $\theta_2\approx$ 72, 96, and 120 hours.
Though the DRAM marginal posterior did not include a peak at 120 hours, this matches the discrete nature of the parameter (as 120 hours would refer to 5 days).
Considering the pairwise scatter plot between the time of initial seeding of infection and the rate of exposure in Figure~\ref{fig:pairwise_compare} for SVI demonstrates the same, though extended, positive correlation as is observed for DRAM.
Further, the correlation between the rate of exposure and the probability of stay at home in Figure~\ref{fig:pairwise_compare} is consistent between the SVI and DRAM posteriors, though again the SVI correlation is extended due to the larger spread of the SVI posterior samples.
The other parameters have a decidedly uniform appearance for the SVI marginal posteriors which contrasts the multi-peaked nature of the DRAM posteriors.
However, the large uncertainty in these parameters is consistent with the GP surrogates lack of sensitivity to them as observed in Table~\ref{tab:feature_importance}.

\begin{figure}
    \centering
    \includegraphics[width=0.8\linewidth]{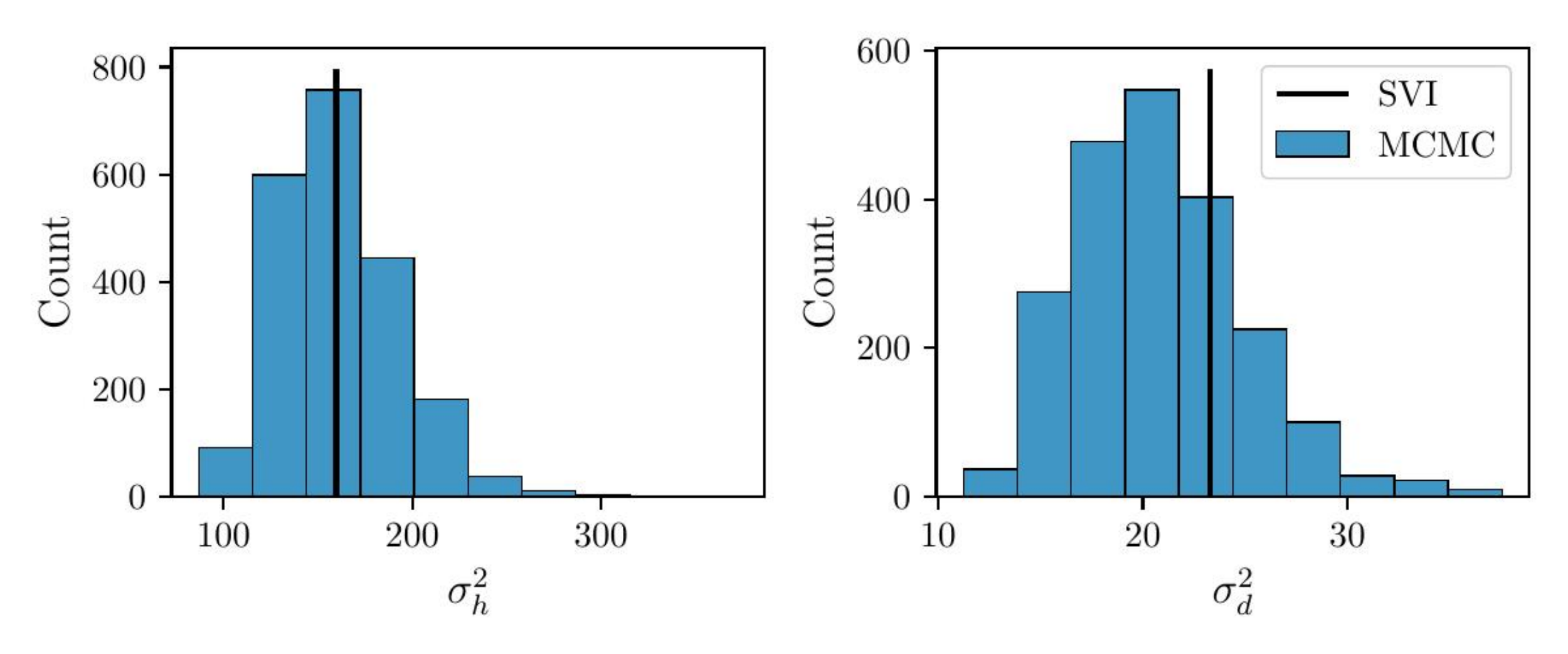}
    \caption{Comparison of the fixed variances values $\sigma_h^2$ and $\sigma_d^2$ used in SVI with the distribution of estimated values from DRAM.}
    \label{fig:variance_comparison}
\end{figure}

Though the qualitative difference in the uncertainty of the DRAM and SVI posteriors may initially appear concerning, it should be noted that the likelihood variances $\sigma_h,\sigma_d$ are fixed for SVI but estimated for DRAM.
A comparison of the uncertainty profiles for these variances computed in DRAM with those fixed for SVI can be seen in Figure~\ref{fig:variance_comparison}.
Reducing the value of $\sigma_h,\sigma_d$ allows for tighter posteriors and more multi-modal behavior.
Given these parameters were estimated in DRAM, it was able to make more use of this flexibility.
To correct for this behavior, some experiments were done to test more complex likelihood formulations for SVI which incorporate time dependent values of $\sigma_h,\sigma_d$ and thus allow for closer replication of the multi-modal DRAM posteriors.
However, the experiments are not included in this text as they are a departure from direct comparison between SVI and DRAM with identical posterior formulations.

\begin{figure}
    \centering
    \includegraphics[width=.9\linewidth]{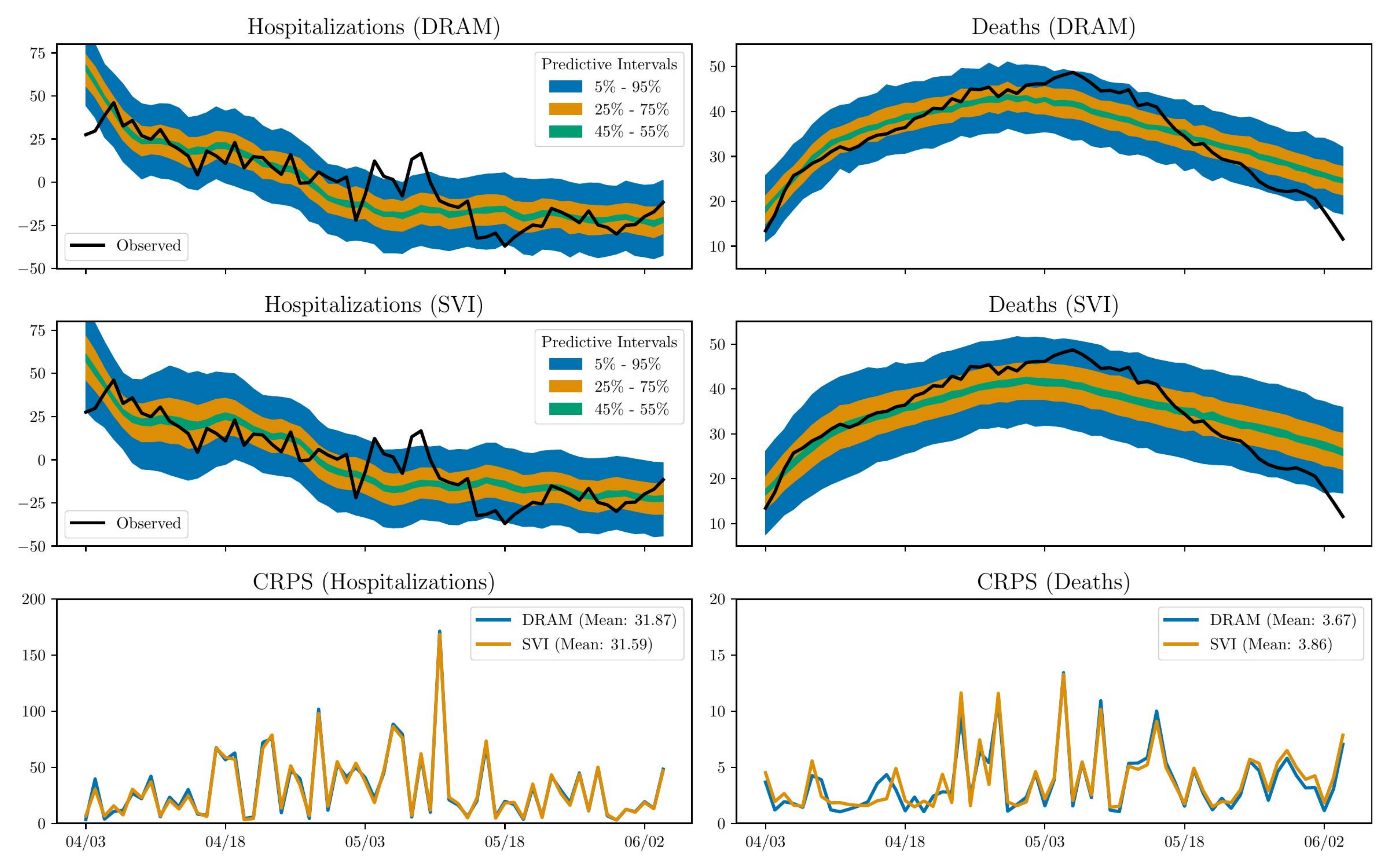}
    \caption{Comparison of posterior predictive distributions using DRAM and SVI calibrations and the GP surrogate. CRPS comparison gives a quantitative comparison of the respective posterior predictive distributions.}
    \label{fig:compare_posterior_predictive}
\end{figure}

In addition to comparing the posterior distributions yielded by the DRAM and SVI calibrations, it is also instructive to consider the predictive performance of these posteriors when pushed through both the GP surrogate and CityCOVID itself.
Figure~\ref{fig:compare_posterior_predictive} illustrates a side-by-side comparison of the predictive posterior distributions of hospitalizations and deaths when using the GP surrogate.
In this figure, the error bounds for the DRAM prediction incorporate the $\sigma_h$ and $\sigma_d$ variances estimated during the calibration and SVI incorporates the fixed values of $\sigma_h$ and $\sigma_d$ as shown in Figure~\ref{fig:variance_comparison}.
The comparison in Figure~\ref{fig:compare_posterior_predictive} also includes the CRPS values across time for hospitalizations and deaths as defined in Equation~\ref{eq:crps}.

\begin{figure}
    \centering
    \includegraphics[width=.9\linewidth]{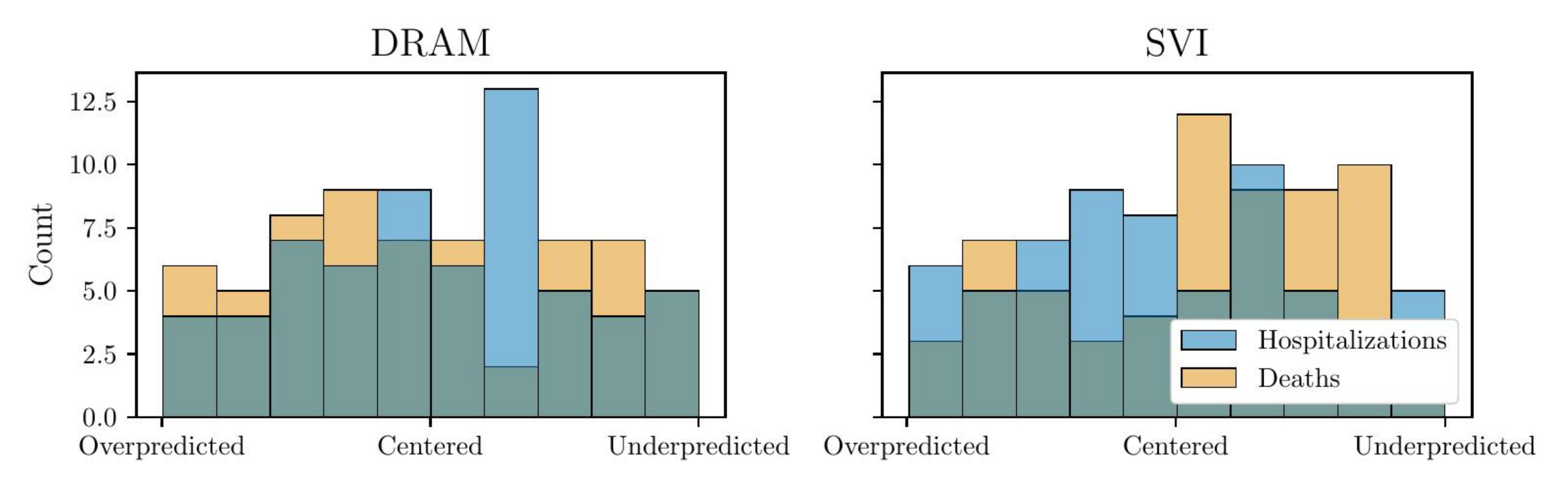}
    \caption{Comparison of verification rank histograms using DRAM and SVI posterior predictive distributions.}
    \label{fig:compare_vrh}
\end{figure}

Observe that the two predictive estimates generally follow similar qualitative trends with the DRAM estimation being slightly tighter around the observed values.
Yet, even with this slight difference between the posterior predictives, the CRPS comparison shows almost no difference in their quantitative accuracy.
To explain this behavior, note that although the SVI predictions have wider uncertainty bounds, it seems also slightly more centered in some key regions (also see Figure~\ref{fig:compare_vrh} for comparison of centering using verification rank histograms).
As CRPS accounts for both the precision and accuracy of the prediction, this centering avoids the score being overly penalized by the increased variance.

\begin{figure}
    \centering
    \includegraphics[width=.9\linewidth]{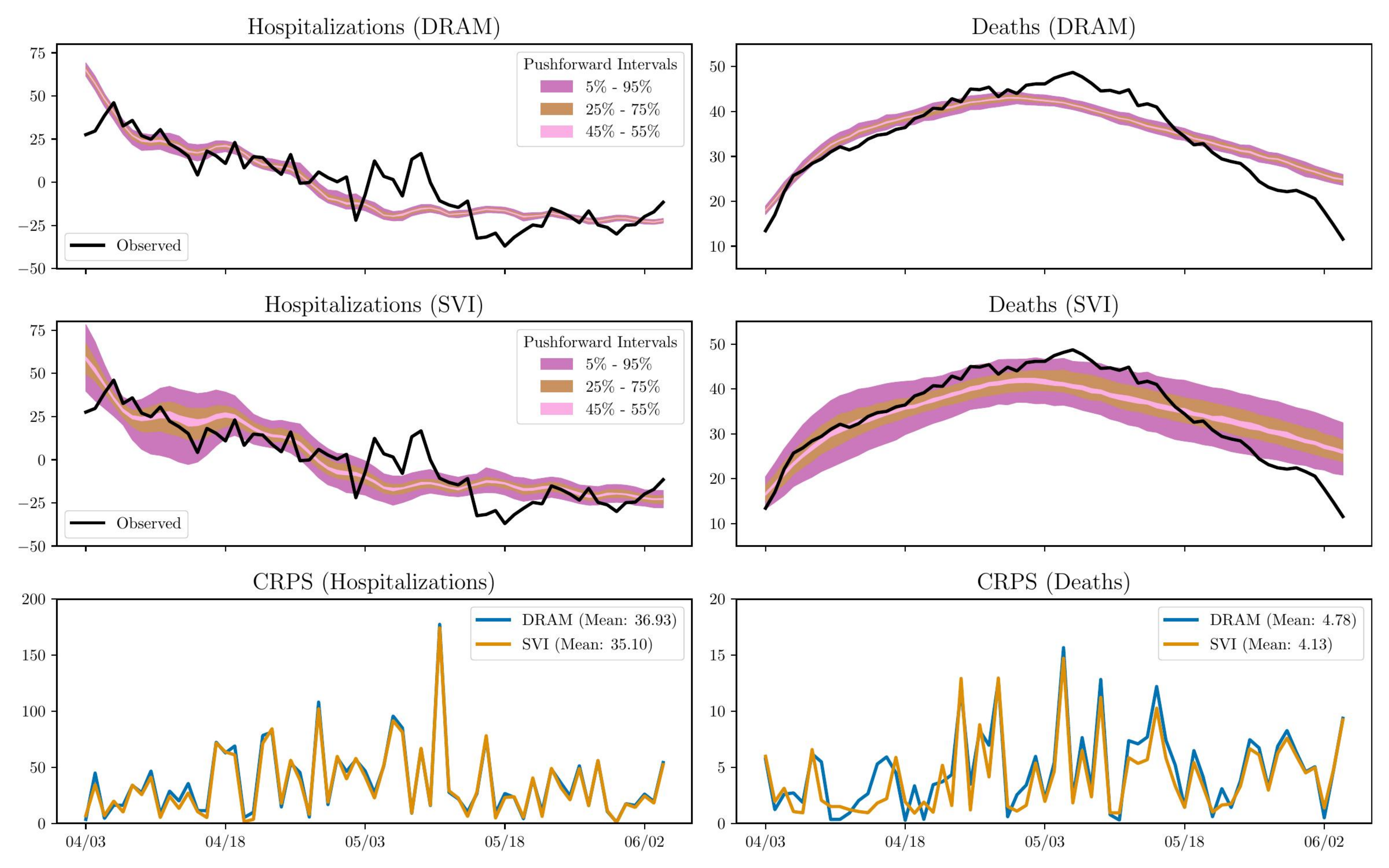}
    \caption{Comparison of posterior pushforward distributions using DRAM and SVI calibrated posteriors and CityCOVID. The CRPS comparison gives a quantitative comparison of the respective posterior pushforward distributions.}
    \label{fig:compare_posterior_pushforward}
\end{figure}

Further comparison can be had by looking at the predictive pushforward distribution yielded when pushing samples from the calibrated DRAM and SVI posteriors through CityCOVID.
A similar side by side comparison of these pushforwards with the observed data is shown in Figure~\ref{fig:compare_posterior_pushforward}.
Given the wider nature of the posteriors yielded by the SVI procedure, the predictions shown in Figure~\ref{fig:compare_posterior_pushforward} are also noticeably more uncertain.
However, in this case, the uncertainty is founded given the noisy nature of the observed hospitalizations and deaths in Chicago.
Quantitatively, this advantage can be seen in the CRPS values which show that SVI provides slightly improved predictive performance given its balance of precision and accuracy.

Generally, the SVI calibration demonstrates predictive performance which rivals  the demonstrated DRAM calibration with GP surrogate or the previously demonstrated DRAM calibration with a random forest surrogate in~\cite{robertson2024_rf, robertson2024_rf_sm} as seen in Figure~\ref{fig:compare_posterior_pushforward}.

\section{Conclusion}
\label{sec:conclusion}

In this paper, we investigated the use of Stein Variational Inference (SVI) as a calibration technique for agent-based models (ABMs) as compared to Bayesian inference, with a focus on the CityCOVID epidemiological model. Our key findings indicate that SVI, when coupled with Gaussian process (GP) surrogates, demonstrated predictive accuracy and calibration effectiveness comparable to traditional Markov Chain Monte Carlo (MCMC) methods. Importantly, SVI offers potential advantages in scalability as discussed in Section~\ref{sec:ip}, which makes it a promising alternative for calibrating high-dimensional ABMs.

While the MCMC approach produced significantly multimodal posterior distributions, reflecting the discrete nature of certain model parameters, SVI yielded more unimodal marginal distributions.
However, SVI did demonstrate the same pairwise correlations and similar modes as the MCMC.
Note that because SVI leverages gradient information in its particle updates, it requires careful hyperparameter tuning and monitoring.
We determined these hyperparameters via experimentation (for learning rate and number of optimization iterations) and self-consistency (for number of particles in SVI).
MCMC, on the other hand, is well-established and robust but can be computationally expensive, particularly for high-dimensional parameter spaces.

After overcoming the challenges of hyperparameter tuning, this work demonstrates that SVI can provide comparable posterior and posterior predictive distributions to Bayesian inference via MCMC.
Additionally, the SVI calibration provided slightly more centered predictions as observed in the verification rank histogram.
As both calibration techniques require significant sampling and thus must make use of a surrogate, SVI provides an excellent approach for scaling to higher dimensional parameter spaces with minimal downside in calibration performance.

The practical application of SVI in stochastic epidemiological models presents several challenges, primarily related to hyperparameter tuning. Effective calibration with SVI necessitates careful selection and adjustment of hyperparameters to ensure convergence and accuracy. This aspect is critical for achieving reliable calibration results and underscores the importance of understanding the underlying model dynamics.

Future directions of this work include demonstrating the efficiency of SVI in the context of high-dimensional ABMs. Specifically, if the parameter space of CityCOVID is not reduced as in~\cite{robertson2024_rf,robertson2024_rf_sm}, can SVI still produce accurate posteriors at acceptable computational expense? This may include developing more robust hyperparameter tuning strategies and exploring adaptive methods to enhance convergence. Further, we plan to explore the use of SVI for a broader range of stochastic ABMs to validate its generalizability and effectiveness across different contexts.

In conclusion, SVI presents a scalable alternative to traditional MCMC methods for the calibration of complex ABMs.
This work lays the groundwork for more efficient and scalable calibration techniques, contributing to the advancement of epidemiological ABMs and their application to public health.

\section*{Acknowledgments}
This material is based upon work supported by the U.S. Department of Energy, Office of Science, Office of Advanced Scientific Computing Research. This paper describes objective technical results and analysis. Any subjective views or opinions that might be expressed in the paper do not necessarily represent the views of the U.S. Department of Energy or the United States Government. This article has been authored by an employee of National Technology \& Engineering Solutions of Sandia, LLC under Contract No. DE-NA0003525 with the U.S. Department of Energy (DOE). The employee owns all right, title and interest in and to the article and is solely responsible for its contents. The United States Government retains and the publisher, by accepting the article for publication, acknowledges that the United States Government retains a non-exclusive, paid-up, irrevocable, world-wide license to publish or reproduce the published form of this article or allow others to do so, for United States Government purposes. The DOE will provide public access to these results of federally sponsored research in accordance with the DOE Public Access Plan https://www.energy.gov/downloads/doe-public-access-plan.  This material is based upon work supported by the National Science Foundation under Grant 2200234, the U.S. Department of Energy, Office of Science, under contract number DE-AC02-06CH11357 and the Bio-preparedness Research Virtual Environment (BRaVE) initiative. This research was completed with resources provided by the Laboratory Computing Resource Center at Argonne National Laboratory.

\bibliographystyle{plain}
\bibliography{references}

\end{document}